\documentclass[runningheads]{llncs}
\usepackage{graphicx}
\usepackage{siunitx}
\usepackage{amsmath}
\usepackage{amssymb}
\usepackage{refcount}
\usepackage{caption}
\usepackage{multirow}
\begin{document}

\title{Semi-supervised Instance Segmentation with a Learned Shape Prior
\thanks{This work was supported by the Deutsche Forschungsgemeinschaft (Research Training Group 2416 MultiSenses-MultiScales).}}

%
%

\author{Long Chen\inst{1}\orcidID{0000-0002-5280-4727} \and
Weiwen Zhang\inst{1} \and
Yuli Wu \inst{1} \and
Martin Strauch\inst{1}\orcidID{0000-0001-6754-211X} \and
Dorit Merhof\inst{1}\orcidID{0000-0002-1672-2185}}

\authorrunning{L. Chen et al.}
%
\institute{Institute of Imaging \& Computer Vision, RWTH Aachen University, Germany\\
\email{\{long.chen, martin.strauch, dorit.merhof\}@lfb.rwth-aachen.de}\\
\url{https://www.lfb.rwth-aachen.de/}}

\maketitle              
\begin{abstract}
To date, most instance segmentation approaches are based on supervised learning that requires a considerable amount of annotated object contours as training ground truth. Here, we propose a framework that searches for the target object based on a shape prior. The shape prior model is learned with a variational autoencoder that requires only a very limited amount of training data: In our experiments, a few dozens of object shape patches from the target dataset, as well as purely synthetic shapes, were sufficient to achieve results en par with supervised methods with full access to training data on two out of three cell segmentation datasets. Our method with a synthetic shape prior was superior to pre-trained supervised models with access to limited domain-specific training data on all three datasets. Since the learning of prior models requires shape patches, whether real or synthetic data, we call this framework semi-supervised learning. The code is available to the public\footnote{https://github.com/looooongChen/shape\_prior\_seg}.  

\keywords{Semi-supervised \and Instance segmentation \and Shape prior \and Variational autoencoder \and Edge loss}
\end{abstract}

\section{Introduction}
\label{sec:introduction}
Instance segmentation, where many instances of an object have to be segmented in one image, is the basis of several practically relevant applications of computer vision, such as cell tracking~\cite{cellTracking}. Many approaches~\cite{mrcnn,stardist,embInstance} have been proposed for instance segmentation, the majority of which are based on supervised learning. The practical applicability of these methods is often limited by the lack of a large training dataset with manually outlined objects. Here, we introduce an instance segmentation approach that only relies on a shape prior which can be learned from a considerably smaller number of training samples or even synthetic data.

The shape is one of the most informative cues in object segmentation and detection tasks. Anatomically constrained neural networks (ACNNs)~\cite{acnn} improve segmentation results by including a shape prior for model regularization. For segmentation refinement, a shape prior has been used by~\cite{shapePost} as a separate post-processing step. Segmentations generated by the shape prior model are reconstructed to the original MRI images through several convolutional layers in~\cite{anatoShape}. By minimizing the reconstruction error, the segmentation model can be trained in an unsupervised fashion. All these works report promising results, but are limited to cases where object position and extent are roughly the same in all images, such as for the cardiac images in~\cite{acnn}, the lung X-ray images in~\cite{shapePost} and the brain MRI scans in~\cite{anatoShape}. To our knowledge, this is the first work considering instance segmentation based on a shape prior, i.e.\ we detect and segment multiple, scattered object instances. Similar to~\cite{spair}, we use the spatial transformer~\cite{transformer} to localize objects. The main advantage of using the spatial transformer lies in its differentiability, making the whole framework end-to-end trainable.

The main contributions of this work are: We propose (1) an semi-supervised instance segmentation approach that seaches for target objects based a shape prior, and (2) a novel loss computing the difference between two gradient maps. This framework provides a way to achieve instance segmentation with a small amount of manual annotations, or by utilizing unpaired annotations (where the correspondence between annotations and images is unknown). We compared our approach to the state-of-the-art supervised method, Mask R-CNN~\cite{mrcnn}, in different training scenarios. On three experimental datasets, our approach is proved to be en par with a Mask R-CNN with full access to training data, while it outperforms a pre-trained Mask R-CNN with limited access to domain-specific training data.

\section{Approach}
\label{sec:approach}
\begin{figure}[!t]
	\centering
	
	\includegraphics[width=\textwidth]{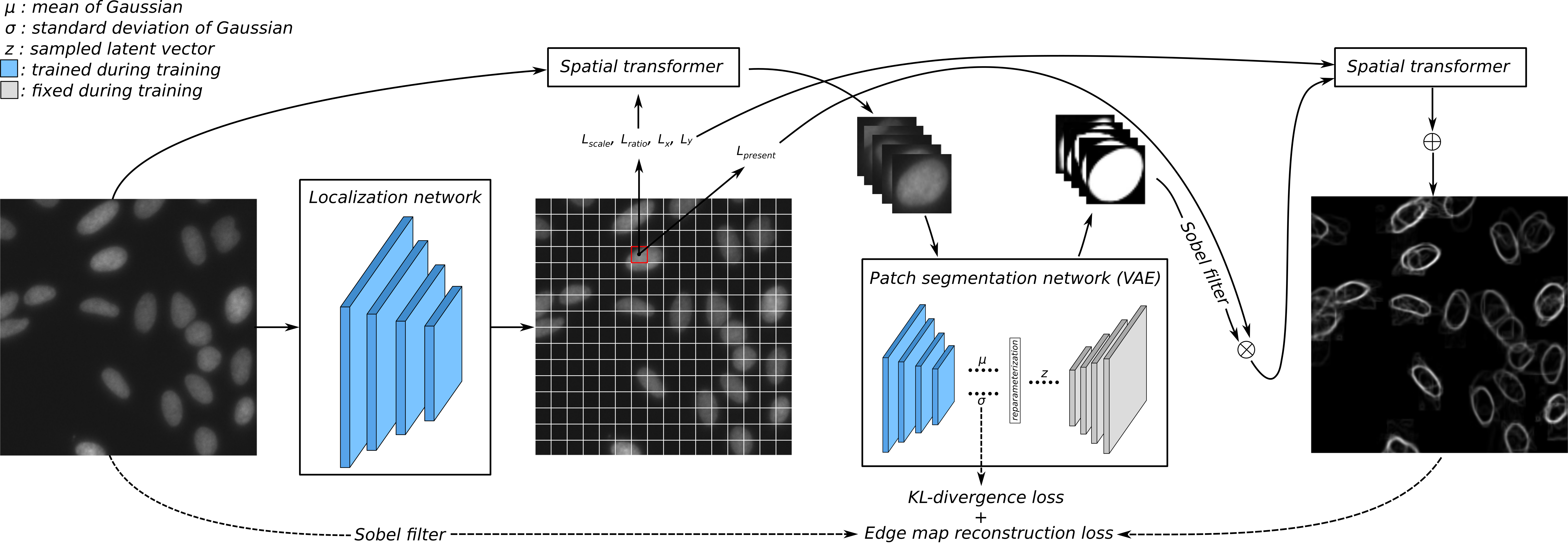}
	
	\caption{Architecture of our framework: the localization network predicts the object position and a presence score, based on which object patches are cropped by a spatial transformer. A variational autoencoder with the decoder part fixed (shape prior) is responsible for the patch segmentation. At last, the gradient maps of segmented patches are stitched together. The model is trained by minimizing the reconstruction loss of the gradient map with the KL-divergence loss as regularization.}
	\label{fig:network}
	\vspace{-3em}
\end{figure}

As shown in Figure~\ref{fig:network}, our framework consists of three main parts: 1) the localization network, 2) the spatial transformer~\cite{transformer}, and 3) the patch segmentation network. Based on the localization prediction, the spatial transformer crops local patches and feeds them to the patch segmentation network. The gradient maps of segmented patches are then stitched together. The entire model is trained by minimizing the reconstruction error of the gradient map. 

During training, the model learns to predict the object position and to find the correspondence between the image patch and the segmentation. The shape prior model (gray part in Fig.~\ref{fig:network}; fixed during training) is guaranteed to output a plausible shape, but the correspondence has to be learned by the model itself.

\subsection{Localization network}
\label{subsec:localization}

The localization network consists of 8 convolutional layers and 4 max pooling layers after every 2 convolutional layers. Given an image of size $(H_{img}, W_{img})$, the localization network will spatially divide the image into an $(H_{img}/S_{cell}, W_{img}/S_{cell})$ grid of cells, where $S_{cell}$ is the cell size and also the downsampling rate. Since 4 pooling layers with stride 2 are used, we have $S_{cell}=16$.

\begin{figure}[]
	\centering
	
	\includegraphics[width=\textwidth]{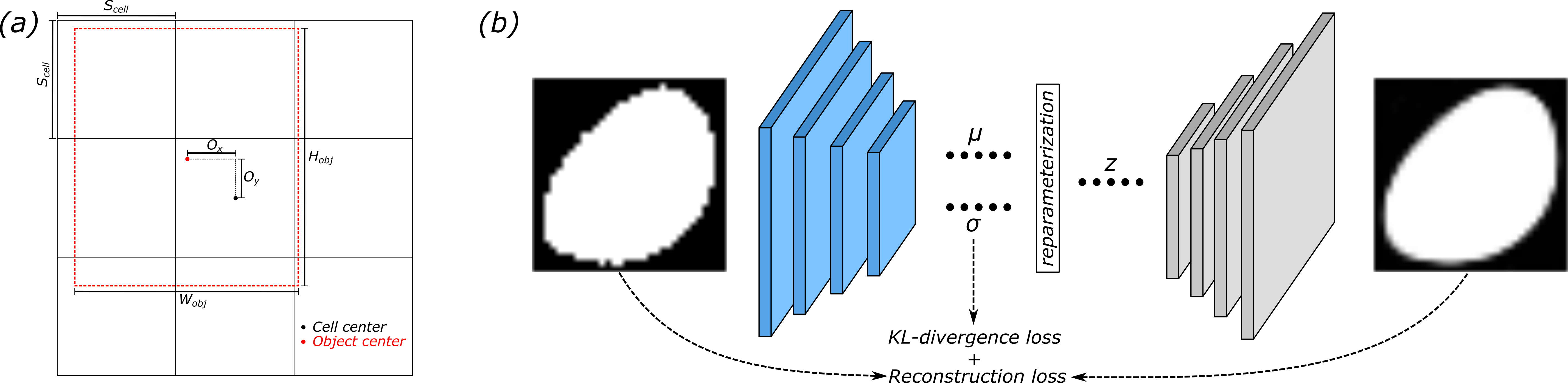}
	
	\caption{(a) Demonstration of parameters of a bounding box. (b) Architecture of the patch segmentation network, which is firstly trained with shape patches. During the detector training, the decoder part is fixed and plays the role of shape prior. }
	\label{fig:box+prior}
	\vspace{-2em}
\end{figure}

Each cell is responsible to predict the presence of an object $L_{presence} \in [0, 1]$, its range described by the bounding box size $(H_{obj}, W_{obj})$ and the offset with respect to the cell center $(O_x, O_y)$ (Figure~\ref{fig:box+prior}(a)), with the implementation:

\begin{align*}
L_{presence} =&\: sigmoid(f_{presence})\\
L_{scale} =&\: sigmoid(f_{scale})\cdot(S_{max}-S_{min}) + S_{min}\\
L_{ratio} =&\: \exp(tanh(f_{ratio}) \cdot \log (R_{max}))\\
(L_{x}, L_{y}) =&\: (0.5 \cdot tanh(f_x), 0.5\cdot tanh(f_y)) 
\end{align*}

\noindent where $f_{[\cdot]}$ is the corresponding input feature map. $sigmoid(\cdot)$ and $tanh(\cdot)$ denote the sigmoid and tanh activation function. $S_{min}$, $S_{max}$ and $R_{max}$ are hyperparameters, which are the minimal scale, the maximal scale and the maximal aspect ratio, respectively. The position is parameterized according to:

\vspace{-1.5em}

\begin{align*}
(H_{obj}, W_{obj}) =&\: (L_{scale} \cdot S_{cell} / \sqrt{L_{ratio}},\: L_{scale} \cdot S_{cell} \cdot \sqrt{L_{ratio}}) \\
(O_x, O_y) =&\: (L_x \cdot S_{cell},\: L_y \cdot S_{cell})
\end{align*}

It is worth mentioning that the maximal offset is $0.5\cdot S_{cell}$, which means that an object will be detected by the cell in which its center lies. 

\subsection{Patch crop and stitch}
Given the location parameters obtained from the localization network, we use a spatial transformer to crop local patches. The spatial transformer implements the crop by sampling transformed grid points, which is differentiable, enabling end-to-end training. The patch crop of the \textit{i}-th cell can be described by transform:

\vspace{-1.2em}
\begin{align*}
	T_{crop}^i
	&=
	\begin{bmatrix}
	W_{img}/W_{obj}^i & 0 & W_{img}\cdot(X_{cell}^i+O_{y}^i)/W_{obj}^i\\
	0 & H_{img}/H_{obj}^i & H_{img}\cdot(Y_{cell}^i+O_{x}^i)/H_{obj}^i\\
	0 & 0 & 1
	\end{bmatrix}
\end{align*}
\vspace{-1em}

\noindent where $(X_{cell}^i, Y_{cell}^i)$ is the cell center. $(O_{x}^i, O_{y}^i)$ and $(H_{obj}^i, W_{obj}^i)$ are the predicted offset and size of the object. All cropped patches will be rescaled to size $S_{patch} \times S_{patch}$ ($S_{patch}=32$ in this work) and segmented by the patch segmentation network, as described in Section~\ref{subsec:shapePrior}. After that, the gradient map of segmented objects will be stitched together by adding up back transformed patches through:

\vspace{-1.2em}
\begin{align*}
T_{stitch}^i
&=
\begin{bmatrix}
W_{obj}^i/S_{patch} & 0 & X_{cell}^i+O_{y}^i\\
0 & H_{obj}^i/S_{patch} & Y_{cell}^i+O_{x}^i\\
0 & 0 & 1
\end{bmatrix}
\end{align*}
\vspace{-1em}

The gradient map is computed by applying the x- and y-directional Sobel filter to the image and taking the square root of the summed square. The gradient map is normalized to range 0 to 1. In this work, we use an input size of $256\times256$ for all experiments. Considering $S_{cell}=16$, 256 patches are cropped in total.

\vspace{-0.5em}
\subsection{Shape prior and patch segmentation network}
\label{subsec:shapePrior}

Similar to~\cite{acnn,shapePost,anatoShape}, we employ a variational autoencoder (VAE) as our shape model. As shown in Figure~\ref{fig:box+prior}(b), the model is trained to reconstruct plausible patch segmentation masks with the KL-divergence loss as regularization.Compared to a standard autoencoder, a VAE learns a more continuous latent space, which is expected to generate plausible new shapes that do not appear in training data.  

In this work, the VAE is trained with $32\times32$ patches. The encoder and decoder consist of 6 convolutional layers and 3 pooling/upsampling layers, respectively. Based on our experiments, model training requires only a small amount of data, especially when the shape variation is small. We train the shape prior with either annotations from a single image or synthetic data (Section~\ref{sec:results}).

After training, the decoder part will be used as the shape prior in the detector (Figure~\ref{fig:network}). Its parameters will be fixed during the detector training. The encoder will be reinitialized and trained together with the localization network.

\vspace{-0.5em}
\subsection{Training}
\label{subsec:training}
The model is trained end-to-end by minimizing the gradient map reconstruction error with the KL-divergence loss as regularization. In initial experiments, we found the mean absolute/squared error (MAE/MSE) to be very unstable during training: The shape prior model tends to generate distorted shapes or degenerates into empty output. Thus, we propose the following novel loss:

\begin{equation}
	L_{edge} = 1-\frac{\frac{1}{N}\sum_{i}min^2(G_{image}^i, G_{reconstruction}^i)}{\frac{1}{N}\sum_{i}G_{reconstruction}^i + \alpha}
\end{equation}

\noindent where $G_{image}$ and $G_{reconstruction}$ indicate the gradient map of the image and the reconstructed gradient map. $N$ is the number of pixels. The $min()$ operation are conducted pixelwise. The parameter $\alpha$ prevents the model from pushing $G_{reconstruction}$ to zero and is set to 0.01 empirically.

Instead of optimizing the value of each pixel, as MSE and MAE, this loss maximizes the proportion of the reconstructed gradient map under the image gradient map. In addition, the square operator in the numerator is proved to be crucial for stable training in our experiments. Our interpretation is that the square operator modulates the back-propagated gradient with the reconstructed gradient map, giving more emphasis to positions around the edge. 

\vspace{-0.5em}
\subsection{Pre- and post-processing}
\label{subsec:postprocessing}
To reduce the influence of extreme values on the loss, we equalized the image and the gradient map by clipping and streching. For all datasets, we truncated the gradient map at 0.8 times the maximum and normalized the value to the range 0 to 1. In addition, we also performed image equalization for the Fluo-N2DH-SIM+ dataset due to the bright spots inside the cell (Figure~\ref{fig:results}). The clip value was set to $1.2$ times the image mean. 

As post-processing, we first filtered out predictions with $L_{presence}$ smaller than 0.1. Non-max suppression is then performed to eliminate duplicate predictions: An instance mask is compared with another mask, when the overlapping area is larger than $p_{non\_max}=0.1$ with respect to its own area. A mask is only retained if its score is the highest in all comparisons.

\section{Experiments and results}
\label{sec:results}

\subsection{Datasets and experiments}
\label{subsec:dataset}
We evaluate our approach on three datasets: the BBBC006 dataset\footnote{\label{ds_bbbc}https://data.broadinstitute.org/bbbc} and two datasets Fluo-N2DH-SIM+ and PhC-C2DL-PSC from the cell tracking challenge~\cite{cellTracking}. In the following, we use BBBC, FLUO and PHC as abbreviations. The BBBC dataset contains 768 microscopic images of human U2OS cells, while the FLUO (HL60 cells with Hoechst staining) and PHC (pancreatic stem cells on a polystyrene substrate) datasets are smaller with 215 and 202 annotated images.

For comparison, we also report the performance of the supervised method Mask R-CNN. The following experiments are performed:

\noindent\textbf{Ours-annotation:} We first evaluate our approach with the shape prior learned from manual annotations. We only took segmentation patches from one image. Specifically, 67, 8 and 138 object patch masks were used for the BBBC, FLUO and PHC shape model training. To model small shape changes and object rotation, we performed rotation (in steps of 30 degrees) and elastic deformation~\cite{elas} to augment the training set. The scale range and maximal aspect ratio was set to 2-3/3, 1-2/1.5 and 1-2/3, respectively.  

\noindent\textbf{Ours-synthetic:} Since the objects are approximately circular, especially for the BBBC and FLUO datasets, we could train the shape prior model with synthetic data consisting of elastically deformed ellipses~\cite{elas} with random angle and major-minor axis ratio. The maximal major-minor axis ratio was 2, 1.5 and 3 for the BBBC, FLUO and PHC dataset, respectively.  
 
\noindent\textbf{MRCNN-scratch-one/full:} We trained a Mask R-CNN from scratch using ResNet-50 backbone. The anchor box scale, aspect ratio and non-maximum suppression (NMS) threshold were set to values equivalent to those used in our approach. Since the Ours-annotation scenario can be considered as one image training, we also trained a Mask R-CNN with one image for comparison.

\noindent\textbf{MRCNN-finetune-one/full:} Since the dataset in our experiments is small, especially for FLUO and PHC, we pretrained the Mask R-CNN on the MS COCO dataset\footnote{https://cocodataset.org/}. Afterwards, we finetuned the model,
with only the head layers trainable, on the actual target dataset. 

For the BBBC and PHC dataset, we cropped images to $256\times256$ and $128\times128$ for training and test. All images were resized to $256\times256$ for the network input. For the scenarios using one training image (Ours-annotation, MRCNN-scratch-one, MRCNN-finetune-one), the images a01\_s1, 02/t000, 02/t150 were used for BBBC, FLUO and PHC, respectively. MRCNN-scratch-full and MRCNN-finetune-full used a01\_s1-b24\_s2, 02/t000-t149, 02/t150-t250 for training. Ours-synthetic requires no manual annotations. All remaining images were kept for testing. 

\vspace{-0.5em}
\subsection{Results and discussion}
\label{subsec:results}

We report the \textit{average precision}\footnote{\label{ds_dsb}https://www.kaggle.com/c/data-science-bowl-2018} (AP) over a range of IoU (intersection over union) thresholds from 0.3 to 0.9 as the evaluation score (Table~\ref{tab:results}). Our approach, including the evaluation scenarios where the shape prior is learned from one image annotation and synthetic data, outperforms the Mask R-CNN trained or finetuned with one image, which shows the advantage of our approach in cases where few or no annotations are available. Furthermore, our approach achieves comparable results with the Mask R-CNN trained/finetuned with the full training set on the BBBC and FLUO dataset, while the performance gap is apparent for the PHC dataset. 

While Mask R-CNN achieved the best mean AP (mAP) on the BBBC dataset, our approach outperformed Mask R-CNN on the FLUO dataset by a relatively large margin. The main reason is that the FLUO dataset is indeed a very small one for Mask R-CNN training, even with finetuning. This again illustrates the advantage of our method on small datasets.

\begin{figure}[t!]
	\centering
	
	\begin{minipage}[b]{0.17\textwidth}
		\centering
		Ground Truth 
		\includegraphics[width=\linewidth]{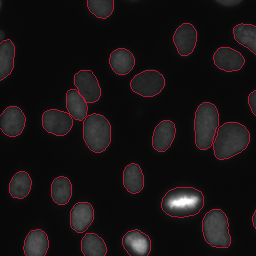} 
	\end{minipage}
	\begin{minipage}[b]{0.17\textwidth}
		\centering
		Ours-synthetic
		\includegraphics[width=\linewidth]{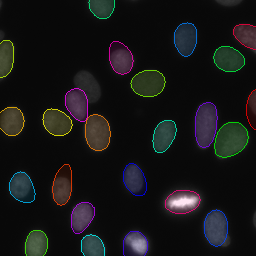} 
	\end{minipage} 
	\begin{minipage}[b]{0.17\textwidth}
		\centering
		Ours-annotation
		\includegraphics[width=\linewidth]{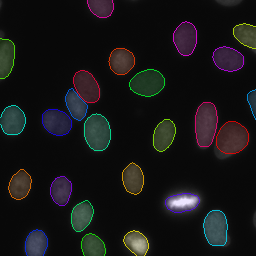} 
	\end{minipage} 
	\begin{minipage}[b]{0.17\textwidth}
		\centering
		MRCNN-finetune-one
		\includegraphics[width=\linewidth]{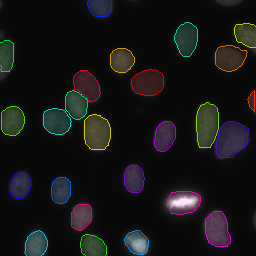} 
	\end{minipage} 
	\begin{minipage}[b]{0.17\textwidth}
		\centering
		MRCNN-finetune-full
		\includegraphics[width=\linewidth]{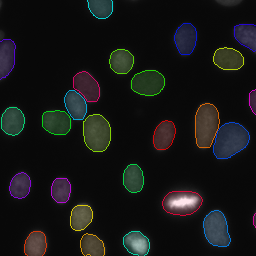} 
	\end{minipage}  
	
	\includegraphics[width=.17\textwidth]{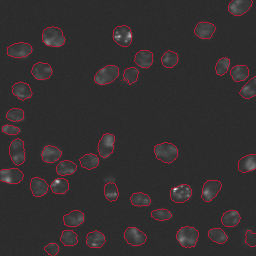} 
	\includegraphics[width=.17\textwidth]{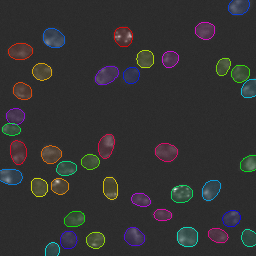} 
	\includegraphics[width=.17\textwidth]{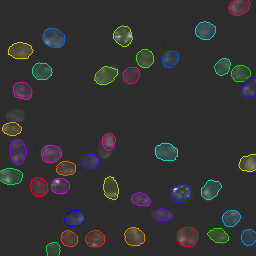} 
	\includegraphics[width=.17\textwidth]{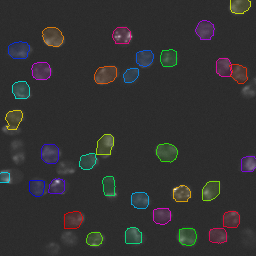} 
	\includegraphics[width=.17\textwidth]{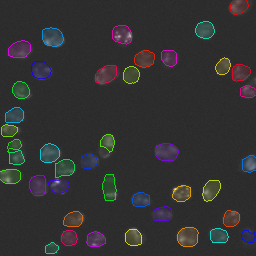} 
	
	\includegraphics[width=.17\textwidth]{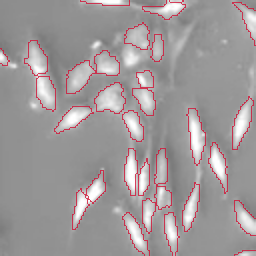} 
	\includegraphics[width=.17\textwidth]{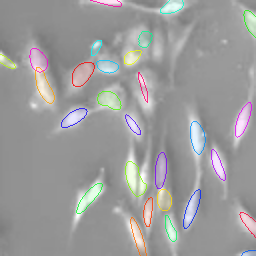} 
	\includegraphics[width=.17\textwidth]{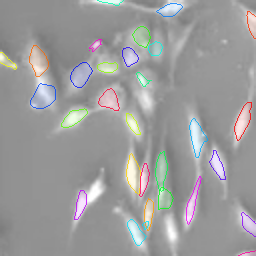} 
	\includegraphics[width=.17\textwidth]{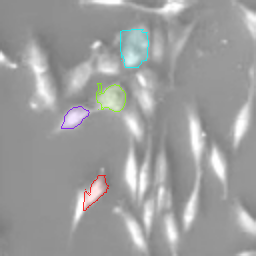} 
	\includegraphics[width=.17\textwidth]{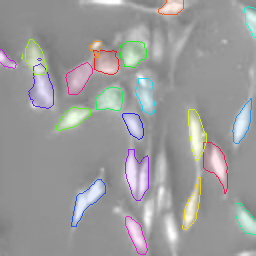} 
	
	\caption{Qualitative results: from top to bottom, the rows show the results on the BBBC006,  Fluo-N2DH-SIM+ and PhC-C2DL-PSC datasets, respectively.}
	\label{fig:results}
	\vspace{-0.5cm}
\end{figure}

On the PHC dataset, neither method performed particularly well. Both methods tended to detect nearby objects as one if there was no clearly visible edge between them. The average precision of our method in the low IoU range was close to or better than that of Mask R-CNN. Figure~\ref{fig:results} shows that our method could detect most objects as well as the Mask R-CNN. However, our method has been designed to heavily rely on the edge clue, so that the segmentation will converge to strong edges. For the PHC dataset, the object boundaries do not generally correspond to the strongest edges. This explains why objects were undersegmented by our approach (Figure~\ref{fig:results}) and why the average precision decreased rapidly with increasing IoU~(Table~\ref{tab:results}).  

The performance improvement through training the shape prior with manually outlined shapes depends on the nature of the shape. On the FLUO dataset, annotated data and synthetic data shape priors performed almost equally well, while training with manual annotations was superior on the other two datasets, even though only a few dozen shapes were used. 

\begin{table}[!t]
	\centering
	\caption{Average precision ($AP$) over different $IoU$ for different datasets (the best two scores in bold). Experiments and abbreviations are introduced in Section~\ref{subsec:dataset}.}
	\label{tab:results}
	\begin{tabular}{c|c|cccccccc}
		\hline
		 Dataset & $IoU$ & 0.3 & 0.4 & 0.5 & 0.6  & 0.7  &  0.8 & 0.9 & $\text{m}AP$ \\
		\hline
		\multirow{6}{*}{BBBC} & Ours-annotation & \textbf{.8345} & \textbf{.8260} & \textbf{.7977} & \textbf{.7632} & .7083 & .6100 & .2660  & \textbf{.6865} \\
		& Ours-synthetic & .8171 & .8012 & .7641 & .7170 & .6525 & .5247 & .2042 & .6401 \\
		& MRCNN-scratch-one & .6386 & .5934 & .5459 & .4769 & .3543 & .1759 & .0294 & .4020 \\
		& MRCNN-scratch-full & .7901 & .7851 & .7708 & .7473 & \textbf{.7128} & \textbf{.6296} & \textbf{.3374} & .6817  \\
		& MRCNN-finetune-one & .7672 & .7524 & .7277 & .7020 & .6608 & .5492 & .1250 & .6121 \\
		& MRCNN-finetune-full & \textbf{.7997} & \textbf{.7949} & \textbf{.7851} & \textbf{.7720} & \textbf{.7521} & \textbf{.6923} & \textbf{.3485} & \textbf{.7064}  \\
		\hline
		\multirow{6}{*}{FLUO} & Ours-annotation & \textbf{.9605} & \textbf{.9538} & \textbf{.9312} & \textbf{.8999} & \textbf{.8228} & \textbf{.6777} & \textbf{.1332} & \textbf{.7685} \\
		& Ours-synthetic & \textbf{.9600} & \textbf{.9497} & \textbf{.9336}  & \textbf{.8986} & \textbf{.8324} & \textbf{.6768} & \textbf{.1378} & \textbf{.7698} \\
		& MRCNN-scratch-one & .0458 & .0324 & .0156 & .0018 & .0000 & .0000 & .0000 &  .0014 \\
		& MRCNN-scratch-full & .9333 & .9144 & .8703 & .7605 & .5765 & .2556 & .01073 & .6173\\
		& MRCNN-finetune-one & .8224 & .8133 & .7905 & .7389 & .5909 & .2404 & .0049 & .5716  \\
		& MRCNN-finetune-full & .9361 & .9252 & .8955 & .8467 & .7265 & .4115 & .0197 & .6802 \\
		\hline
		\multirow{6}{*}{PHC} & Ours-annotation & \textbf{.6840} & \textbf{.6034} & .4035 & .1468 & .0233 & .0028 & .0000 & .2662  \\
		& Ours-synthetic & .6471 & .5611 & .3605 & .1326 & .0219 & .0027 & .0000 & .2466 \\
		& MRCNN-scratch-one & .1124 & .0991 & .0847 & .0668 & .0353 & .0049 & .0000 & .0576  \\
		& MRCNN-scratch-full & .6332 & .6001 & \textbf{.5226} & \textbf{.4467} & \textbf{.2981} & \textbf{.1079} & \textbf{.0023} & \textbf{.3730}  \\
		& MRCNN-finetune-one & .1647 & .1602 & .1460 & .1146 & .0633 & .0108 & .0000 & .0942  \\
		& MRCNN-finetune-full & \textbf{.6551} & \textbf{.6380} & \textbf{.5855} & \textbf{.5014} & \textbf{.3425} & \textbf{.1144} & \textbf{.0007} & \textbf{.4053}  \\
		\hline
	\end{tabular}
	\vspace{-1em}
\end{table}

\vspace{-1em}
\section{Conclusion and outlook}
\label{sec:conclusion}


We have proposed an instance segmentation framework which searches for target objects in images based on a shape prior model. In practice, this allows segmenting instances with a very limited amount of annotations, segmenting synthesizable shapes without any annotation, as well as reusing object annotations from other datasets.

The main limitation of our approach lies in the dependency on the edge cues. Images should have a relatively clear background, which is, however, the case for many biomedical datasets\footnotemark[\getrefnumber{ds_dsb}]. Future work will focus on including area-based information, which will make our approach applicable to further datasets, e.g.\ in cases where edges and object boundaries do not always coincide.

%

%
%
%
%

\end{document}